\documentclass[11pt]{article}
\usepackage{lingmacros,
            fullname,
	    tree-dvips,
	    latexsym }

\makeatletter
\def\downlett#1{$_{\hbox{\tiny\rm #1}}$}

\newcommand{\exno}[1]{\setcounter{tempcnt}{\value{enums}}%
\addtocounter{tempcnt}{#1}%
\arabic{tempcnt}}

\def\section{\@startsection {section}{1}{\z@}{-3.5ex plus -1ex minus
 -.2ex}{2.3ex plus .2ex}{\normalsize\bf}}
\def\subsection{\@startsection{subsection}{2}{\z@}{-3.25ex plus -1ex minus
 -.2ex}{1.5ex plus .2ex}{\normalsize\bf}}

\def \pred {\mbox{\sc pred}}
\def \subj {\mbox{\sc subj}}
\def \obj {\mbox{\sc obj}}
\def \spec {\mbox{\sc spec}}

\newcommand{\lam}[1]{\lambda #1. }

\newcommand{\linimp}{\;\mbox{$-\hspace*{-.4ex}\circ$}\;}
\newcommand{\means}{\makebox[1.2em]{$\leadsto$}}
\newcommand{\meansub}[1]{\makebox[1.5em]{$\leadsto_{#1}$}}
\newcommand{\IT}[1]{\mbox{\it #1\/}}
\newcommand{\BF}[1]{\mbox{\bf #1}}
\newcommand{\All}[1]{\forall #1.\;}
\newcommand{\attr}[2]{\mbox{$(#1\;#2)$}}
\newcommand{\pex}[1]{(\ref{#1})}

\def\fd#1{\setlength{\baselineskip}{0pt}
  \small
     \hbox{#1}}
\def\fdx#1{\setlength{\baselineskip}{0pt}\vcenter{#1}}
\def\fdand#1{\(\left[\fdx{#1}\right]\)}
\def\feat#1#2{\vskip
.4ex\hbox{\hspace{.2em}#1\hspace{1em}#2\hspace{.2em}}\vskip .8ex}
\def\fdset#1{\(\left\{\hbox{#1}\right\}\)}

\makeatother

\title{Linear Logic for Meaning Assembly\thanks{
Dalrymple, Lamping, Saraswat: Xerox PARC, Palo Alto CA 94304;
\{dalrymple,lamping,saraswat\}@parc.xerox.com.  Pereira: AT\&T Bell
Laboratories, Murray Hill NJ 07974; pereira@research.att.com}}
\author{Mary Dalrymple\\
John Lamping\\
Fernando Pereira\\
Vijay Saraswat}

\begin{document}
\maketitle

\section{Introduction}

Semantic theories of natural language associate meanings with
utterances by providing meanings for lexical items and rules for
determining the meaning of larger units given the meanings of their
parts.  Meanings are often assumed to combine via function
application, which works well when constituent structure trees are
used to guide semantic composition.  However, we believe that the {\em
functional structure}\/ of Lexical-Functional Grammar is best used to
provide the syntactic information necessary for constraining
derivations of meaning in a cross-linguistically uniform format.  It
has been difficult, however, to reconcile this approach with the
combination of meanings by function application.

In contrast to compositional approaches, we present a deductive
approach to assembling meanings, based on reasoning with constraints,
which meshes well with the unordered nature of information in the
functional structure.  Our use of {\it linear logic} as a `glue' for
assembling meanings allows for a coherent treatment of the LFG
requirements of completeness and coherence as well as of modification
and quantification.

\section{The Framework}

This paper provides a brief overview of our ongoing investigation in
the use of formal deduction to explicate the relationship between
syntactic analyses in Lexical-Functional Grammar (LFG) and semantic
interpretations
\cite{DLS:EACL,DHLS:ROCLING,DLPS:Intensional,DLPS:Quantification,DKLS:Coord,DLPS:QuantLFG}.  We
use {\em linear logic\/} \cite{Girard:Linear} to represent the
connection between two dissimilar linguistic levels: LFG f-structures
and their semantic interpretations.

F-structures provide a uniform representation of syntactic information
relevant to semantic interpretation that abstracts away from the
varying details of phrase structure and linear order in particular
languages.  But as noted by Halvorsen \shortcite{Halvorsen:LI} and
Reyle \shortcite{Reyle:CompSem}, the flatter, unordered functional
structure of LFG does not fit well with traditional semantic
compositionality, based on functional abstraction and application,
which mandates a rigid order of semantic composition.  We are thus led
to a more relaxed form of compositionality, in which, as in more
traditional ones, the semantics of each lexical entry in a sentence is
used exactly once in interpretation, but without imposing a rigid
order of composition. Approaches to semantic interpretation that
encode semantic representations in attribute-value structures
\cite{PollardSag:HPSG1,FenstadEtAl:SitLgLogic,PollardSag:HPSG2} offer
such a relaxation of compositionality, but are unable to properly
represent constraints on variable binding and scope
\cite{Pereira:SemComp}.

The present approach, in which linear logic is used to specify the
relation between f-structures and their meanings, provides exactly
what is required for a calculus of semantic composition for LFG. It
can directly represent the constraints on the creation and use of
semantic units in sentence interpretation, including those pertaining
to variable binding and scope, without forcing a particular
hierarchical order of composition, except as required by the
properties of particular lexical entries.

\subsection{Syntax}

In languages like English, the substantial scaffolding provided by
surface constituent structure trees is often a useful guide for
semantic composition, and the $\lambda$-calculus is a convenient
formalism for assembling the semantics along that scaffolding
\cite{Montague:Formal}.  This is because the derivation of the meaning
of a phrase can often be viewed as mirroring the surface constituent
structure of the English phrase.  The sentence `Bill appointed
Hillary'\/ has the surface constituent structure indicated by the
bracketing in \ex{1}:
\enumsentence{[\downlett{S} [\downlett{NP} Bill] [\downlett{VP} appointed
[\downlett{NP} Hillary]]]}
The verb is viewed as bearing a close syntactic relation to the object
and forming a constituent with it; this constituent then combines with
the subject of the sentence.  Similarly, the meaning of the verb can
be viewed as a two-place function which is applied first to the
object, then to the subject, producing the meaning of the sentence.

However, this approach is not as natural for languages whose surface
structure does not resemble English.  For instance, a problem is
presented by VSO languages such as Irish
\cite{McCloskey:SynSem}.   To preserve the hypothesis that surface
constituent structure provides the proper scaffolding for semantic
interpretation in VSO languages, one of two assumptions must be made.
One must assume either that semantic composition is nonuniform across
languages (leading to loss of explanatory power), or that semantic
composition proceeds not with reference to surface syntactic
structure, but instead with reference to a more abstract
(English-like) constituent structure representation.  This second
hypothesis seems to us to render vacuous the claim that surface
constituent structure is useful in semantic composition.

Further problems are encountered in the semantic analysis of a free
word order language such as Warlpiri \cite{Simpson:Warlpiri}, where
surface constituent structure does not always give rise to units that
are semantically coherent or useful. Here, an argument of a verb may
not even appear as a single unit at surface constituent structure;
further, arguments of a verb may appear in various different places in
the string. In such cases, the appeal to an order of composition
different from that of English is particularly unattractive, since
different orders of composition would be needed for each possible word
order sequence.

The observation that surface constituent structure does not always
provide the optimal set of constituents or hierarchical structure to
guide semantic interpretation has led to efforts to use a more
abstract structure to guide semantic composition.  As originally
proposed by Kaplan and Bresnan \shortcite{KaplanBresnan:LFG} and
Halvorsen \shortcite{Halvorsen:LI}, the {\em functional structure}\/
or {\em f-structure}\/ of LFG is a representation of such a structure.
The c-structure and f-structure for sentence \pex{ex:bah} are given in
\pex{ex:bahfs}:

\enumsentence{\label{ex:bah} Bill appointed Hillary.}

\enumsentence{\label{ex:bahfs}\evnup{
\begin{tabular}{cc}
C-structure: & F-structure: \\[1em]
\modsmalltree{3}{\mc{3}{\node{l}{S}\hspace*{1em}}\\
            \node{m}{NP}&\mc{2}{\hspace*{1em}\node{n}{VP}}\\
                        &\node{p}{V}&\node{t}{NP}\\
            \node{q}{Bill}&\node{r}{appointed}&\node{s}{Hillary}}
\nodeconnect{l}{m}
\nodeconnect{l}{n}
\nodeconnect{m}{q}
\nodeconnect{n}{p}
\nodeconnect{n}{t}
\nodeconnect{t}{s}
\nodeconnect{p}{r}
&
\fd{
\fdand{\feat{\pred}{`{\sc appoint}'}
       \feat{\subj}{\fdand{\feat{\pred}{`{\sc Bill}'}}}
       \feat{\obj}{\fdand{\feat{\pred}{`{\sc Hillary}'}}}}}
\end{tabular}}}

\noindent As illustrated, an f-structure represents the more abstract
syntactic function-argument structure of the sentence, encoding
relations such as \subj\ and \obj.  Those relations are realized in
different c-structure forms in different languages, but are
represented directly and uniformly in the f-structure.  Formally, an
f-structure consists of a collection of attributes, such as \pred,
\subj, and \obj, whose values can, in turn, be other f-structures.

We will not provide a detailed account of the relation between
c-structure and f-structure; for such an account, see
\namecite{Bresnan:MentalRep}, \namecite{LRZ:LFG}, and the references
cited there.  Here, we will begin with the f-structures for the
examples we discuss, and concentrate on an exposition of how the
f-structure provides information necessary to carry out a semantic
deduction.

\subsection{Semantic Projections}

Following work by Kaplan \shortcite{Kaplan:3Sed} and Halvorsen and
Kaplan \shortcite{HalvorsenKaplan:Projections}, we make use of a {\em
semantic\/} or $\sigma$-{\em projection\/} function $\sigma$ to map
f-structures to {\em semantic\/} or $\sigma$-{\em structures\/}
encoding information about f-structure meaning. For a particular use
of `Bill', the resulting configuration is:

\vbox{\enumsentence{\evnup{\fd{$h:$\node{a}{\fdand{
       \feat{\pred}{`{\sc Bill}'}}}}
\hspace*{2em}%
\fd{$h_\sigma:$\node{b}{[\ ]}}$\means \IT{Bill}$
\anodecurve[br]{a}[bl]{b}{2em}}}}

\noindent The semantic projection function $\sigma$ is represented by
an arrow.  We use labels such as $h$ to refer to particular
f-structures.  The association between the semantic structure
$h_\sigma$ and a meaning $P$ can be represented by the atomic formula
$h_\sigma \means P$, which we will refer to as a {\it meaning
constructor}, where $\means$ is an otherwise uninterpreted binary
predicate symbol.  (In fact, we use not one but a family of
relations~$\meansub{\tau}$ indexed by the semantic type of the
intended second argument, although for simplicity we will omit the
type subscript whenever it is determinable from context.)  For the
case at hand, if a particular occurrence of `Bill' in a sentence is
associated with f-structure $h$, the semantic constraint will be
instantiated as: $$h_\sigma \means \IT{Bill}$$ representing the
association between $h_\sigma$ and the constant $\IT{Bill}$
representing its meaning.

We will often informally say that $P$ is $h$'s meaning without
referring to the role of the semantic structure $h_\sigma$ in
$h_\sigma \means P$. Actually, however, f-structures and their
semantic projections must be distinguished, because semantic
projections can carry more information than just the association to
the meaning for the corresponding f-structure; see
\namecite{DLPS:Quantification} for more discussion.

As noted above, the $\lambda$-calculus is not a very natural tool for
combining meanings of f-structure constituents.  The problem is that
the subconstituents of an f-structure are unordered, and so the fixed
order of combination of a functor with its arguments imposed by the
$\lambda$-calculus is no longer an advantage; in fact, it becomes a
disadvantage, since an artificial ordering must be imposed on the
composition of meanings. Furthermore, the components of the
f-structure may be not only complements but also modifiers, which
contribute to the final semantics in a very different way.

Instead, we assume that lexical entries contribute premises -- meaning
constructors -- to a logical deduction.  The meaning constructor is a
linear logic formula that can be understood as `instructions' for
combining the meanings of the lexical entry's syntactic arguments to
obtain the meaning of the f-structure headed by the entry.  In effect,
then, our approach uses linear logic as the `glue' with which semantic
representations are assembled.  Once all the constraints are
assembled, deduction in the logic is used to infer the meaning of the
entire structure.  Throughout this process we maintain a sharp
distinction between assertions about the meaning (the glue) and the
meaning itself.

In the case of the verb `appointed', the meaning constructor is a
formula consisting of instructions on how to assemble the meaning of a
sentence with main verb `appointed', given the meanings of its subject
and object.  The verb `appointed' contributes the following
f-structure and meaning constructor:

\vbox{\enumsentence{\evnup{\fd{$f:$\node{a}{\fdand{
       \feat{\pred}{`{\sc appoint}'}
       \feat{\subj}{[\ ]}
       \feat{\obj}{[\ ]}}}}
\hspace*{2em}%
\fd{$f_\sigma:$\node{b}{[\ ]}}}\\[1em]
$\All{ X, Y}\attr{f}{\subj}_\sigma\means X \otimes \attr{f}{\obj}_\sigma\means
Y \linimp
                f_\sigma \means \IT{appoint}\/(X, Y)$
\anodecurve[r]{a}[bl]{b}{2.5em}}}

\noindent The meaning constructor, given in the last line of
(\exno{0}), asserts that:
\begin{itemize}
\item if $f$'s subject $\attr{f}{\subj}$ has
meaning $X$
\item and ($\otimes$) $f$'s object $\attr{f}{\obj}$ has meaning $Y$
\item then ($\linimp$) $f$ has meaning $\IT{appoint}(X,Y)$.
\end{itemize}
The linear-logic connectives of multiplicative conjunction $\otimes$
and linear implication $\linimp$ are used to specify how the meaning
of a clause headed by the verb is composed from the meanings of the
arguments of the verb. For the moment, we can think of the linear
connectives as playing the same role as the analogous classical
connectives conjunction $\wedge$ and implication $\rightarrow$, but we
will see that the specific properties of the linear connectives are
essential to guarantee that lexical entries bring into the
interpretation process all and only the information provided by the
corresponding words.

We will now show how meanings are assembled by linear-logic deduction.
For readability, we will present derivations informally. As a first
example, consider the following f-structures:

\enumsentence{\label{ex:bahafs}\evnup{\fd{
$f$:\fdand{\feat{\pred}{`{\sc appoint}'}
       \feat{\subj}{$g$:\fdand{\feat{\pred}{`{\sc Bill}'}}}
       \feat{\obj}{$h$:\fdand{\feat{\pred}{`{\sc Hillary}'}}}}}}}

\noindent From  the lexical entries for `Bill',
`Hillary', and `appointed', we obtain the following meaning
constructors, abbreviated as \BF{bill}, \BF{hillary}, and
\BF{appointed}:\footnote{For the sake of illustration, we will provide
only the simplest semantics for the lexical entries we discuss,
ignoring (among other issues) the representation of tense and aspect.}
\[
\begin{array}{@{\protect\strut}ll@{\protect\strut}}
\BF{bill}\colon& g_{\sigma} \means \IT{Bill}\\
\BF{hillary}\colon& h_{\sigma} \means \IT{Hillary}\\
\BF{appointed}\colon& \All{ X, Y} g_{\sigma}\means X \otimes h_{\sigma}\means Y
\linimp
                f_{\sigma}\means \IT{appoint}(X, Y)
\end{array}
\]
In the following, assume that the formula \BF{bill-appointed} is
defined thus:
\[\begin{array}[t]{ll}
\BF{bill-appointed}\colon&
\All{ Y}h_{\sigma}\means Y \linimp f_{\sigma} \means \IT{appoint}(\IT{Bill}, Y)
\end{array}
\]
\noindent Then the following derivation is possible in linear logic ($\vdash$
stands
for the linear-logic entailment relation):
\enumsentence{\label{ex:bahderiv}
$
\begin{array}[t]{l@{\hspace*{2em}}ll}
&\BF{bill} \otimes \BF{hillary} \otimes \BF{appointed} & (Premises.) \\[.5ex]
\vdash &  \BF{bill-appointed} \otimes \BF{hillary} & X\mapsto
\IT{Bill}\\[0.5ex]
\vdash & f_{\sigma} \means \IT{appoint}(\IT{Bill}, \IT{Hillary}) & Y\mapsto
\IT{Hillary}
\end{array}
$}
Of course, another derivation is also possible. Assume that the
formula \BF{appointed-hillary} is defined as:
\[\begin{array}[t]{ll}
\BF{appointed-hillary}\colon&
\All{ X}g_{\sigma}\means X \linimp f_{\sigma} \means \IT{appoint}(X,
\IT{Hillary})
\end{array}
\]
\noindent Then we have the following derivation:
\enumsentence{
$
\begin{array}[t]{l@{\hspace*{2em}}ll}
&\BF{bill} \otimes \BF{hillary} \otimes \BF{appointed} & (Premises.) \\[.5ex]
\vdash &  \BF{bill} \otimes \BF{appointed-hillary} & Y\mapsto
\IT{Hillary}\\[0.5ex]
\vdash & f_{\sigma} \means \IT{appoint}(\IT{Bill}, \IT{Hillary}) & X \mapsto
\IT{Bill}
\end{array}
$}
yielding the same conclusion.

In summary, each word in a sentence contributes a linear-logic
formula, its meaning constructor, relating the semantic projections of
specific f-structures in the LFG analysis to representations of their
meanings.  From these formulas, the interpretation process attempts to
deduce an atomic formula relating the semantic projection of the whole
sentence to a representation of the sentence's meaning. Alternative
derivations can sometimes yield different such conclusions,
corresponding to ambiguities of semantic interpretation.

In LFG, syntactic predicate-argument structure is projected from
lexical entries.  Therefore, its effect on semantic composition will
for the most part -- in fact, in all the cases considered in this
paper -- be determined by lexical entries, not by phrase-structure
rules.  In particular, the phrase-structure rules for S and VP in the
examples discussed above need not encode semantic information, but
only specify how grammatical functions such as \subj\ are expressed in
English.  In some cases, the constituent structure of a syntactic
construction may make a direct semantic contribution, as when
properties of the construction as a whole and not just of its lexical
elements are responsible for the interpretation of the
construction. Such cases include, for instance, relative clauses with
no complementizer (`the man Bill met').  We will not discuss
construction-specific interpretation rules in this paper.

\section{Linear logic}

An important motivation for using linear logic is that it allows us to
directly capture the intuition that lexical items and phrases each
contribute exactly once to the meaning of a sentence.  As noted by
\namecite[page~172]{KleinSag:Type}:
\begin{quote}
Translation rules in Montague semantics have the property that the
translation of each component of a complex expression occurs exactly
once in the translation of the whole.  \ldots That is to say, we do
not want the set S [of semantic representations of a phrase] to
contain {\em all\/} meaningful expressions of IL which can be built up
{}from the elements of S, but only those which use each element exactly
once.
\end{quote}

\noindent In our terms, the semantic contributions of the constituents
of a sentence are not context-independent assertions that may be used
or not in the derivation of the meaning of the sentence depending on
the course of the derivation. Instead, the semantic contributions are
{\em occurrences\/} of information which are generated and used exactly
once.  For example, the formula $g_{\sigma}\means
\IT{Bill}$ can be thought of as providing one occurrence of the meaning
$\IT{Bill}$ associated to the semantic projection $g_{\sigma}$.  That
meaning must be consumed exactly once (for example, by \BF{appointed} in
\pex{ex:bahderiv}) in the derivation of a meaning of the entire utterance.

It is this `resource-sensitivity' of natural language semantics---an
expression is used exactly once in a semantic derivation---that linear
logic can model. The basic insight underlying linear logic is that
logical formulas are {\em resources\/} that are produced and consumed
in the deduction process.  This gives rise to a resource-sensitive
notion of implication, the {\em linear implication\/} $\linimp$: the
formula $A \linimp B$ can be thought of as an action that can {\em
consume\/} (one copy of) $A$ to produce (one copy of) $B$. Thus, the
formula $A \otimes (A \linimp B)$ linearly entails $B$.  It does not
entail $A \otimes B$ (because the deduction consumes $A$), and it does
not entail $(A \linimp B) \otimes B$ (because the linear implication
is also consumed in doing the deduction).  This resource-sensitivity
not only disallows arbitrary duplication of formulas, but also
disallows arbitrary deletion of formulas. Thus the linear
multiplicative conjunction $\otimes$ is sensitive to the multiplicity
of formulas: $A \otimes A$ is not equivalent to $A$ (the former has
two copies of the formula $A$).  For example, the formula $A \otimes A
\otimes (A \linimp B)$ linearly entails $A \otimes B$ (there is still
one $A$ left over) but does not entail $B$ (there must still be one
$A$ present).  The following table provides a summary:
$$\begin{array}{lc@{\linimp}c}
\mbox{INCORRECT:} & A & (A \otimes A)\\
\mbox{INCORRECT:} & (A \otimes B) & A\\
\mbox{CORRECT:} & (A \otimes (A \linimp B)) & B\\
\mbox{INCORRECT:} & (A \otimes (A \linimp B)) & (A \otimes B)\\
\mbox{INCORRECT:} & (A \otimes (A \linimp B)) & (A \linimp B) \otimes B
\end{array}$$
In this way, linear logic checks that a formula is used
once and only once in a deduction, enforcing the requirement that each
component of an utterance contributes exactly once to the assembly of
the utterance's meaning.

A direct consequence of the above properties of linear logic is that
the constraints of functional {\em completeness}\/ and {\em coherence}
hold without further stipulation\footnote{`An f-structure is {\it
locally complete}\/ if and only if it contains all the governable
grammatical functions that its predicate governs.  An f-structure is
{\em complete}\/ if and only if all its subsidiary f-structures are
locally complete. An f-structure is {\em locally coherent}\/ if and
only if all the governable grammatical functions that it contains are
governed by a local predicate.  An f-structure is {\em coherent}\/ if
and only if all its subsidiary f-structures are locally coherent.'
\cite[pages~211--212]{KaplanBresnan:LFG}

To illustrate:
\enumsentence[(a)]{*John devoured. [incomplete]}
\enumsentence[(b)]{*John arrived Bill the sink. [incoherent]}}
\cite{DLS:EACL}.  In the present setting, the feature structure $f$
corresponding to the utterance is associated with the ($\otimes$)
conjunction $\phi$ of all the formulas associated with the lexical
items in the utterance. The conjunction is said to be {\em complete\/}
and {\em coherent\/} iff $Th \vdash \phi \linimp f_{\sigma} \means t$
(for some term $t$), where $Th$ is the background theory of general
linguistic principles.  Each possible $t$ is to be thought of as a
valid meaning for the sentence. This guarantees that the entries are
used exactly once in building up the denotation of the utterance: no
syntactic or semantic requirements may be left unfulfilled, and no
meaning may remain unused.

Our glue language needs to be only a fragment of higher-order linear
logic, the {\em tensor fragment}, that is closed under conjunction,
universal quantification, and implication.  This fragment arises from
transferring to linear logic the ideas underlying the concurrent
constraint programming scheme of \namecite{Saraswat:PhD}.\footnote{
\namecite{SaraswatLincoln:HLCC} provide
an explicit formulation for the higher-order version of the linear
concurrent constraint programming scheme. \namecite{Scedrov:Linear}
and \namecite{Troelstra:LinLog} give a tutorial introduction to linear
logic itself; \namecite{Saraswat:IntroLCC} supplies further background on
computational aspects of linear logic relevant to the implementation
of the present proposal.}

Our approach shares a number of commonalities with various systems of
categorial syntax and semantics. In particular, the Lambek calculus
\cite{Lambek:SentStruct}, introduced as a logic of syntactic
combination, turns out to be a fragment of noncommutative
multiplicative linear logic.  For a discussion of how our approach
compares to the approach of Lambek and related approaches
\cite{Moortgat:PhD,Hepple:PhD,Morrill:Intensional}, see
\namecite{DLPS:Quantification}.

\section{Modification}

A primary advantage of the use of linear logic `glue' in the
derivation of meanings of sentences is that it enables a clear
treatment of modification, as described in
\namecite{DLS:EACL}. Consider the following sentence, containing the
sentential modifier `obviously':
\enumsentence{\label{ex:bokh}
Bill obviously appointed Hillary.}
We make the standard assumption that the verb `appointed' is the main
syntactic predicate of this sentence.  The following is the
f-structure for example (\ref{ex:bokh}):
\enumsentence{\evnup{\fd{
$f$:\fdand{\feat{\sc pred}{`kiss'}
       \feat{\sc subj}{$g$:\fdand{\feat{\sc pred}{`Bill'}}}
       \feat{\sc obj}{$h$:\fdand{\feat{\sc pred}{`Hillary'}}}
       \feat{\sc mods}{\fdset{\fdand{\feat{\sc pred}{`obviously'}}}}}}}}
We also assume that the meaning of the sentence can be represented by the
following formula:
\enumsentence{
$obviously(kiss(Bill, Hillary))$}
It is clear that there is a mismatch between the syntactic
representation and the meaning of the sentence; syntactically, the
verb is the main functor, while the main semantic functor is the
adverb.\footnote{The related phenomenon of {\it head switching},
discussed in connection with machine translation by Kaplan et al.\
\shortcite{Kaplan:Translation} and Kaplan and Wedekind
\shortcite{KaplanWedekind:EACL}, is also amenable to treatment along
the lines presented here.}

Recall that linear logic enables a coherent notion of {\it
consumption} and {\it production} of meanings.  We claim that the
semantic function of adverbs (and, indeed, of modifiers in general) is
to consume the meaning of the structure they modify, producing a new,
modified meaning.  Note below that the meaning of the modified
structure $f$ in the meaning constructor contributed by `obviously'
appears on {\it both} sides of $\linimp$; the unmodified meaning is
consumed, and the modified meaning is produced.

The derivation of the meaning of example (\ref{ex:bokh}) is:

\begin{center}
$\begin{array}{@{\protect\strut}lll@{\protect\strut}}
\BF{bill}\colon& g_{\sigma} \means \IT{Bill}\\
\BF{hillary}\colon& h_{\sigma} \means \IT{Hillary}\\
\BF{appointed}\colon& \All{ X, Y} g_{\sigma}\means X \otimes h_{\sigma}\means Y
\linimp
                f_{\sigma}\means \IT{appoint}(X, Y)\\
\BF{obviously}\colon& (\forall P.\ f_{\sigma} = P~\linimp~f_{\sigma} =
obviously(P))\\
\end{array}$\\[1ex]
$\begin{array}{@{\protect\strut}rll@{\protect\strut}}
&\mbox{{\bf bill} $\otimes$ {\bf hillary} $\otimes$ {\bf appointed} $\otimes$
{\bf
obviously}} & \mbox{(Premises.)}\\
\vdash &  \BF{bill-appointed} \otimes \BF{hillary} \otimes \BF{obviously} &
X\mapsto \IT{Bill}\\
\vdash & f_{\sigma} \means \IT{appoint}(\IT{Bill}, \IT{Hillary})
\otimes \BF{obviously} & Y\mapsto \IT{Hillary}\\
\vdash & \IT{obviously}(\IT{appoint}(\IT{Bill}, \IT{Hillary})) & P\mapsto
\IT{appoint}(\IT{Bill}, \IT{Hillary})
\end{array}$
\end{center}
The first part of the derivation is the same as the derivation for the
sentence `Bill appointed Hillary'.  The crucial difference is the
presence of information introduced by `obviously'.  In the last step
in the derivation, the linear implication introduced by `obviously'
consumes the previous value for $f_\sigma$ and produces the new and
final value.

By using linear logic, each step of the derivation keeps track of what
`resources' have been consumed by linear implications.  As mentioned
above, the value for $f_\sigma$ is a meaning for this sentence only if
there is no other information left.  Thus, the derivation could not
stop at the next to last step, because the linear implication
introduced by `obviously' was still left.  The final step provides the
only complete and coherent meaning derivable for the utterance.

\section{Quantification}

Our treatment of quantification
\cite{DLPS:Intensional,DLPS:Quantification,DLPS:QuantLFG},
and in particular of quantifier scope ambiguities and of the
interactions between scope and bound anaphora, follows the analysis of
Pereira
\shortcite{Pereira:SemComp,Pereira:HOD}, but offers in addition a formal
account of the syntax-semantics interface, which was treated only
informally in that earlier work.

To illustrate our analysis of quantification, we will consider the
following sentence:
\enumsentence{\label{ex:simple-quant}
Bill convinced everyone.}
The f-structure for \pex{ex:simple-quant} is:
\enumsentence{\evnup{
\fd{$f$:\fdand{\feat{\pred}{`{\sc convince}'}
           \feat{\subj}{$g$:\fdand{\feat{\pred}{`{\sc Bill}'}}}
           \feat{\obj}{$h$:\fdand{\feat{\pred}{`{\sc everyone}'}}}}}}}

\noindent We assume that this example has a meaning representable as:
$$\IT{every}(\IT{person}, \lam{z}\IT{convince}(\IT{Bill}, z)) $$
Here, we will work with the meaning of a quantifier like `everyone';
for a full exposition of our analysis of quantification and of how a
determiner like `every' combines with a noun like `person', see
\namecite{DLPS:Quantification}.  The quantifier `everyone' can be seen
as making a semantic contribution along the following lines:

\enumsentence{\label{ex:everyone-unscoped}\evnup{
$\All{ S}  (\All{ x}  f_\sigma \means x \linimp \IT{scope} \means S(x))
\linimp  \IT{scope} \means \IT{every}(\IT{person}, S)$}}

\noindent Informally, the constructor for `everyone' can be read as
follows: if by giving the arbitrary meaning $x$ of type $e$ to $f$,
the f-structure for `everyone', we can derive the meaning $S(x)$ of
type $t$ for the scope of quantification {\em scope}\/, then $S$ can
be the property that the quantifier requires as its scope, yielding
the meaning $\IT{every}(\IT{person}, S)$ for {\em scope}. The
quantified NP can thus be seen as providing two contributions to an
interpretation: locally, a {\em referential import\/} $x$, which must
be discharged when the scope of quantification is established; and
globally, a {\em quantificational import\/} of type
\mbox{$(e\rightarrow t)\rightarrow t$}, which is applied to the
meaning of the scope of quantification to obtain a quantified
proposition.  Notice that the assignment of a meaning to {\em scope}\/
appears on both sides of the implication, as in the case of modifiers,
and that the meaning is not the same in the two instances.

We use the place-holder $\IT{scope}$ in (\ref{ex:everyone-unscoped})
to represent possible choices for the scope of the quantifier, but we
did not specify how this scope was chosen.  Previous work on scope
determination in LFG \cite{HalvorsenKaplan:Projections} defined
possible scopes at the f-structure level, using {\em inside-out
functional uncertainty}\/ to nondeterministically choose a scope
f-structure for quantified noun phrases.  That approach requires the
scope of a quantified NP to be an f-structure which contains the NP
f-structure.  In contrast, our approach depends only on the logical
form of semantic constructors to yield exactly the appropriate scope
choices. Within the constraints imposed by that logical form, the
actual scope can be freely chosen; the linear implication guarantees
that the scope will contain the quantified NP, since only scope
meanings which are obtained by consuming the variable representing the
quantified NP can be chosen.

Logically, this means that the semantic constructor for an NP
quantifies universally over semantic projections of possible scopes,
as follows:
\enumsentence{\label{ex:everyone}
\evnup{$\BF{everyone}\colon\
\All{H, S}  (\All{x}  f_\sigma \means x \linimp H \means S(x))
\linimp  H \means \IT{every}(\IT{person}, S)$}}

The premises for the derivation of the meaning of example
(\ref{ex:simple-quant}) are the meaning constructors for `Bill',
`convinced', and `everyone': $$\begin{array}{ll@{\,}l}
\BF{bill}\colon & \makebox[1em][l]{$g_{\sigma} \means Bill$}\\[1ex]
\BF{convinced}\colon & \All{ X, Y} & g_{\sigma} \means X \otimes h_{\sigma}
\means Y\linimp f_{\sigma} \means convince\/(X, Y)\\[1ex]
\BF{everyone}\colon\ & \All{H,S} &
\/(\All{x}  h_{\sigma} \means x \linimp H \means_t S(x))\\
 & \hfill \linimp & H \means_t \IT{every}(\IT{person}, S)
\end{array}$$
Notice that we have explicitly indicated that the scope of the
quantifier must be of type $t$, by means of the subscript $t$ on the
`means' relation $\means_t$.  This typing is implicit in the schematic
formula for quantifiers given in \pex{ex:everyone-unscoped}.

Giving the name \BF{bill-convinced} to the formula
\[\begin{array}[t]{ll}
\BF{bill-convinced}\colon& \All{Y} h_{\sigma} \means Y \linimp  f_{\sigma}
\means \IT{convince}(\IT{Bill}, Y) \\
\end{array}
\]
we have the following derivation:
\[
\begin{array}{l@{\hspace*{1em}}ll}
&\BF{bill} \otimes \BF{convinced} \otimes \BF{everyone} & \mbox{\/(Premises.)}
\\[0.5ex]
\vdash & \BF{bill-convinced} \otimes \BF{everyone} & X \mapsto
\IT{Bill}\\[0.5ex]
\vdash & f_{\sigma} \means \IT{every}(\IT{person},
\lam{z}\IT{convince}(\IT{Bill}, z)) & H \mapsto f_{\sigma}, Y  \mapsto x \\
& & S \mapsto \lam{z}\IT{convince}(\IT{Bill}, z)
\end{array}
\]
The formula \BF{bill-convinced} represents the semantics of the scope
of the determiner `every'. No derivation of a different formula
$f_{\sigma} \means_t P$ is possible.

While the formula
\[\All{ Y}h_{\sigma} \meansub{e} Y \linimp h_{\sigma} \meansub{e} Y\]
could at first sight be considered another possible scope, the type
subscripting of the \means\ relation used in the determiner lexical
entry requires the scope to represent a dependency of a proposition on
an individual.  But this formula represents the dependency of an
individual on an individual (itself). Therefore, it does not provide a
valid scope for the quantifier.

\section{Quantifier scope ambiguities}

When a sentence contains more than one quantifier, scope ambiguities
are of course possible. In our system, those ambiguities will appear
as alternative successful derivations. We will take as our example
this sentence:\footnote{To allow for apparent scope
ambiguities, we adopt a scoping analysis of indefinites, as proposed,
for example, by \namecite{Neale:Descriptions}.}

\enumsentence{\label{ae}
Every candidate appointed a manager.}

\noindent The f-structure for sentence (\ref{ae}) is:

\enumsentence{\evnup{
\fd{$f$:\fdand{\feat{\pred}{`{\sc appoint}'}
           \feat{\subj}{$g$:\fdand{\feat{\spec}{`{\sc every}'}
                                     \feat{\pred}{`{\sc candidate}'}}}
           \feat{\obj}{$h$:\fdand{\feat{\spec}{`{\sc a}'}
                                     \feat{\pred}{`{\sc manager}'}}}}}}}

\noindent The meaning constructors for `every candidate' and `a manager' are
analogous to the one for `everyone' in the previous section. The
derivation proceeds from those contributions together with the
contribution of `appointed':
$$\begin{array}{lr@{\,}l}
\BF{every-candidate}\colon&
\All{G, R} &(\All{x}  g_{\sigma}\means x \linimp G \means R(x))\\
& \linimp & G \means \IT{every}(\IT{candidate}, R)\\[1ex]
\BF{a-manager}\colon& \All{H,S} & (\All{y}  h_{\sigma} \means y\linimp H \means
S(y))\\
& \linimp & H \means \IT{a}(\IT{manager}, S)\\[1ex]
\BF{appointed}\colon& \All{X, Y} & g_{\sigma} \means X\otimes
h_{\sigma} \means Y \linimp f_{\sigma} \means \IT{appoint}(X, Y)
\end{array}$$

\noindent As of yet, we have not made any commitment about the scopes
of the quantifiers; the scope and scope meaning variables in
\BF{every-candidate} and \BF{a-manager} have not been instantiated.
Scope ambiguities are manifested in two different ways in our system:
through the choice of different semantic structures $G$ and $H$,
corresponding to different scopes for the quantified NPs,
or through different relative orders of application for quantifiers that
scope at the same point.  For this example, the second case is
relevant, and we must now make a choice to proceed. The two possible
choices correspond to two equivalent rewritings of {\bf appointed}:
\[
\begin{array}{ll}
\BF{appointed}_1\colon & \All{ X} g_{\sigma} \means X \linimp (\All{
Y}h_{\sigma} \means Y \linimp f_{\sigma} \means \IT{appoint}(X, Y)) \\
\BF{appointed}_2\colon & \All{ Y}h_{\sigma} \means Y\linimp (\All{ X}
g_{\sigma} \means X \linimp f_{\sigma} \means \IT{appoint}(X, Y))
\end{array}
\]
\noindent These two equivalent forms correspond to the two possible
ways of `currying' a two-argument function $f: \alpha\times
\beta\rightarrow \gamma$ as one-argument functions: $$\lambda u.\lambda
v.f(u,v): \alpha \rightarrow (\beta \rightarrow \gamma)$$
$$\lambda v.\lambda
u.f(u,v): \beta \rightarrow (\alpha \rightarrow \gamma)$$
We select `a manager' to take narrower scope by using the
variable instantiations
\[H\mapsto f_\sigma, Y\mapsto y, S\mapsto \lam{v}\IT{appoint}(X,v)\]
and transitivity of implication to combine
\BF{appointed}$_1$ with {\bf a-manager} into:
$$\begin{array}{@{\protect\strut}lr@{\,}l@{\protect\strut}}
\BF{appointed-a-manager}\colon&
 \All{X} & g_{\sigma}\means X \\
& \linimp & f_{\sigma} \meansub{t} a(\IT{manager}, \lam{v}\IT{appoint}(X, v))
\end{array}$$
This gives the derivation
\[
\begin{array}{l@{\hspace*{2em}}l}
& \BF{every-candidate} \otimes \BF{appointed}_1 \otimes \BF{a-manager}\\[0.5ex]
\vdash &  \BF{every-candidate} \otimes \BF{appointed-a-manager}\\[0.5ex]
\vdash & f_{\sigma} \meansub{t}\IT{every}(\IT{candidate},
\lam{u}
\IT{a}(\IT{manager},\lam{v}\IT{appoint}(u, v)))
\end{array}
\]
of the $\forall\exists$ reading of \pex{ae}, where the last step
uses the substitutions
\[G\mapsto f_{\sigma}, X\mapsto x, R \mapsto \lam{u}
\IT{a}(\IT{manager},\lam{v}\IT{appoint}(u, v))\]

Alternatively, we could have chosen `every candidate' to take narrow
scope, by combining \BF{appointed}$_2$ with {\bf every-candidate} to
produce:
$$\begin{array}{@{\protect\strut}lr@{\,}l@{\protect\strut}}
\BF{every-candidate-appointed}\colon&
 \All{Y} & h_{\sigma}\means Y \\
& \linimp & f_{\sigma} \meansub{t} \IT{every}(\IT{candidate},
\lam{u}\IT{appoint}(u, Y))
\end{array}$$
\noindent This gives the derivation
$$
\begin{array}{l@{\hspace*{2em}}l}
& \BF{every-candidate} \otimes \BF{appointed}_2 \otimes \BF{a-manager}\\[0.5ex]
\vdash &  \BF{every-candidate-appointed} \otimes \BF{a-manager}\\[0.5ex]
\vdash & f_{\sigma}
\meansub{t}\IT{a}(\IT{manager},\lam{v}\IT{every}\/(\IT{candidate},\lam{u}\IT{appoint}(u, v)))
\end{array}
$$ for the $\exists\forall$ reading. These are the only two possible
outcomes of the derivation of a meaning for \pex{ae}, as required
\cite{DLPS:Quantification}.

\section{Conclusion}

Our approach exploits the f-structure of LFG for syntactic information
needed to guide semantic composition, and also exploits the
resource-sensitive properties of linear logic to express the semantic
composition requirements of natural language.  The use of linear logic
as the glue language in a deductive semantic framework allows a
natural treatment of the syntax-semantics interface which
automatically gives the right results for completeness and coherence
constraints and for modification \cite{DLS:EACL}, covers quantifier
scoping and bound anaphora \cite{DLPS:Quantification,DLPS:QuantLFG}
and their interactions with intensionality \cite{DLPS:Intensional},
offers a clean and natural treatment of complex predicate formation
\cite{DHLS:ROCLING}, and extends nicely to an analysis of the
semantics of structure sharing \cite{DKLS:Coord}.

\section{Acknowledgements}

Over the (rather lengthy) course of the development of the approach
outlined here, we have greatly benefited from discussion and
assistance from many of our colleagues; we are particularly grateful
to Andy Kehler, Angie Hinrichs, Ron Kaplan, John Maxwell, Kris
Halvorsen, Joan Bresnan, Stanley Peters, Paula Newman, Chris Manning,
Henriette de Swart, Johan van Benthem, and David Israel for very
helpful comments as well as for incisive criticism.

\end{document}